# Convex Multitask Learning with Flexible Task Clusters

**Leon Wenliang Zhong**                                                    WZHONG@CSE.UST.HK
**James T. Kwok**                                                          JAMESK@CSE.UST.HK
Department of Computer Science and Engineering, Hong Kong University of Science and Technology, Hong Kong

## Abstract

Traditionally, multitask learning (MTL) assumes that all the tasks are related. This can lead to negative transfer when tasks are indeed incoherent. Recently, a number of approaches have been proposed that alleviate this problem by discovering the underlying task clusters or relationships. However, they are limited to modeling these relationships at the task level, which may be restrictive in some applications. In this paper, we propose a novel MTL formulation that captures task relationships at the feature-level. Depending on the interactions among tasks and features, the proposed method construct different task clusters for different features, without even the need of pre-specifying the number of clusters. Computationally, the proposed formulation is strongly convex, and can be efficiently solved by accelerated proximal methods. Experiments are performed on a number of synthetic and real-world data sets. Under various degrees of task relationships, the accuracy of the proposed method is consistently among the best. Moreover, the feature-specific task clusters obtained agree with the known/plausible task structures of the data.

## 1. Introduction

Many real-world problems involve the learning of a number of tasks. Instead of learning them individually, it is now well-known that better generalization performance can be obtained by harnessing the intrinsic task relationships and allowing tasks to borrow strength from each other. In recent years, a number of techniques have been developed under this multitask learning (MTL) framework.

Traditional MTL methods assume that all the tasks are related (Evgeniou & Pontil, 2004; Evgeniou et al., 2005).

However, when this assumption does not hold, the performance can be even worse than single-task learning. If it is known that the tasks are clustered, a simple remedy is to constrain task sharing to be just within the same cluster (Argyriou et al., 2008; Evgeniou et al., 2005). This can be further extended to the case where task relationships are represented in the form of a network (Kato et al., 2007). However, in practice, such an explicit knowledge of task clusters/network is rarely available.

Recently, by adopting different modeling assumptions, a number of approaches have been proposed that identify task relationships simultaneously with parameter learning. For example, some assume that the task parameters share a common prior in a Bayesian model (Yu et al., 2005; Zhang & Schneider, 2010; Zhang & Yeung, 2010); that the data follows a dirty model (Jalali et al., 2010); that most of the tasks lie in a low-dimensional subspace (Ando & Zhang, 2005; Chen et al., 2010), or that outlier tasks are present (Chen et al., 2011a). In this paper, we will mainly be interested in techniques that assume the tasks are clustered (Argyriou et al., 2008; Evgeniou et al., 2005), and then infer the clustering structure automatically during learning (Jacob et al., 2008; Kang et al., 2011). Interestingly, it is recently shown that this clustered MTL approach is equivalent to alternating structure optimization (Ando & Zhang, 2005) that assumes the tasks share a low-dimensional structure (Zhou et al., 2011).

However, all the existing methods model the task relationships at the task level, and the features are assumed to always observe the same set of task clusters or covariance structure (Figure 1(a)). This may be restrictive in some real-world applications. For example, in recommender systems, each customer corresponds to a task and each feature a movie attribute. Suppose that we have a relatively coherent group of customers, such as Jackie Chan fans who are interested in action comedy movies (Figure 1(b)). On the "language" attribute, however, some of them may prefer English, some prefer standard Chinese (Putonghua/Mandarin), some prefer Cantonese or even a combination of these. Hence, the clustering structure as seen by this feature is very different from those of the oth-ers. Another example is when the features are obtained by some feature extraction algorithm (such as PCA) and so have different discrimination abilities. While the less discriminating features may be used in a similar manner by all the tasks, highly discriminating features may be very class-specific and are used differently by different tasks (Figure 1(c)). Hence, again, these features may observe different task relationships. This phenomenon will also be demonstrated in the experiments in Section 3.

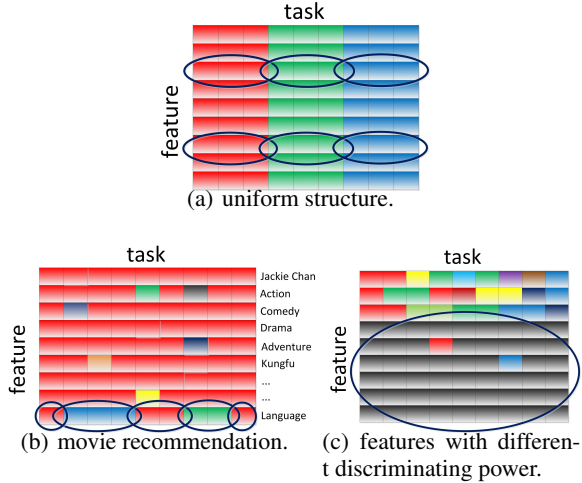

Figure 1. Example task clustering structures of the weight matrix. Each row is a feature, each column is a task, and each color denotes a cluster (colors on different rows are not related). Figure (a) shows an uniform clustering structure shared by all features; while (b) and (c) show examples of non-uniform clustering structures.

In this paper, we extend clustered MTL such that the task cluster structure can vary from feature to feature. This is thus more fine-grained than existing MTL methods that only capture task-level (but not feature-level) relationships. Moreover, a key difference with (Jacob et al., 2008) is that we do not require the number of clusters to be pre-specified. Indeed, depending on the complexity of the tasks and usefulness of each feature, different numbers of clusters can be formed for different features.

Computationally, the optimization problem is often challenging in clustered MTL algorithms. For example, in (Kang et al., 2011), it leads to a mixed integer program, which has to be relaxed as a nonlinear optimization problem and then solved by gradient descent. This suffers from the local minimum problem and potentially slow convergence. On the other hand, the proposed approach directly leads to a (strongly) convex optimization problem, which can then be efficiently solved by accelerated proximal methods (Nesterov, 2007) after some transformations.

**Notation**: Vector/matrix transpose is denoted by the superscript $'$, $\|\mathbf{A}\|_F = \sqrt{\text{trace}(\mathbf{A}'\mathbf{A})}$ is the Frobenius norm of matrix $\mathbf{A}$, $\mathbf{A}_{i\cdot}$ is its $i$th row and $\mathbf{A}_{\cdot j}$ its $j$th column.

## 2. The Model

Suppose that there are $T$ tasks. The $t$th task has $n_t$ training samples $\{(\mathbf{x}_1^{(t)}, y_1^{(t)}), \ldots, (\mathbf{x}_{n_t}^{(t)}, y_{n_t}^{(t)})\}$, with input $\mathbf{x}_i^{(t)} \in \mathbb{R}^D$ and output $y_i^{(t)} \in \mathbb{R}$. We stack the inputs and outputs together to form matrices $\mathbf{X}^{(t)} = [\mathbf{x}_1^{(t)}, \ldots, \mathbf{x}_{n_t}^{(t)}]'$ and $\mathbf{y}^{(t)} = [y_1^{(t)}, \ldots, y_{n_t}^{(t)}]'$, respectively. A linear model is used to learn each task. Let the weight associated with task $t$ be $\mathbf{w}_t$. The predictions on the $n_t$ samples are stored in the vector $\mathbf{X}^{(t)}\mathbf{w}_t$.

### 2.1. Simultaneous Clustering of Task Parameters

We decompose each $\mathbf{w}_t$ into $\mathbf{u}_t + \mathbf{v}_t$, where $\mathbf{u}_t$ tries to capture the shared clustering structure among task parameters, and $\mathbf{v}_t$ captures variations specific to each task. Learning of $\mathbf{w}_t$'s is performed jointly with the clustering of $\mathbf{u}_t$'s via the following regularized risk minimization problem:

$$\min_{\mathbf{U}, \mathbf{V}} \quad \sum_{t=1}^{T} \|\mathbf{y}^{(t)} - \mathbf{X}^{(t)}(\mathbf{u}_t + \mathbf{v}_t)\|^2 + \lambda_1 \|\mathbf{U}\|_{clus}$$
$$+ \lambda_2 \|\mathbf{U}\|_F^2 + \lambda_3 \|\mathbf{V}\|_F^2, \tag{1}$$

where $\mathbf{U} = [\mathbf{u}_1, \ldots, \mathbf{u}_T]$ and $\mathbf{V} = [\mathbf{v}_1, \ldots, \mathbf{v}_T]$, and $\lambda_1, \lambda_2, \lambda_3$ are regularization parameters. The first term in (1) is the empirical (squared) loss on the training data, and $\|\mathbf{U}\|_{clus}$ is the sum of pairwise differences for elements in each row of $\mathbf{U}$,

$$\|\mathbf{U}\|_{clus} = \sum_{d=1}^{D} \sum_{i<j} |U_{di} - U_{dj}|. \tag{2}$$

For each feature $d$ and each $(\mathbf{u}_i, \mathbf{u}_j)$ pair, the pairwise penalty in $\|\mathbf{U}\|_{clus}$ encourages $U_{di}, U_{dj}$ to be close together, leading to feature-specific task clusters. It can also be shown that $\|\mathbf{U}\|_{clus}$ is a convex relaxation of $k$-means clustering on each feature. Note that this is different from the fused lasso regularizer (Tibshirani et al., 2005), which is used for clustering features in single-task learning while $\|\mathbf{U}\|_{clus}$ is for clustering tasks in MTL. It is also different from the graph-guided fused lasso (GFlasso) (Chen et al., 2011b), which does not decompose $\mathbf{w}_t$ as $\mathbf{u}_t + \mathbf{v}_t$, and subsequently cannot cluster the tasks due to the use of smoothing. The regularizer $\|\mathbf{V}\|_F^2 = \sum_{t=1}^{T} \|\mathbf{v}_t\|^2$ penalizes the deviations of each $\mathbf{w}_t$ from $\mathbf{u}_t$, and $\|\mathbf{U}\|_F^2$ is the usual ridge regularizer penalizing $\mathbf{U}$'s complexity. Since $\|\mathbf{U}\|_F^2, \|\mathbf{V}\|_F^2$ are strongly convex and the other terms in (1) are convex, (1) is a strongly convex optimization problem.

Some MTL papers also decompose $\mathbf{w}_t$ as $\mathbf{u}_t + \mathbf{v}_t$, but the formulations and goals are different from ours. In (Evgeniou et al., 2005), $\mathbf{u}_t$ is the (single) cluster center of all the tasks; in (Ando & Zhang, 2005; Chen et al., 2010; 2011a), $\mathbf{u}_t$ comes from a low-dimensional linear subspace, which is extended to a nonlinear manifold in (Agarwal et al., 2010); in (Jalali et al., 2010), $\mathbf{u}_t$ is the component that uses features shared by other tasks.

Moreover, model (1) encompasses a number of interesting special cases: (i) $\lambda_1 \to \infty$:[1] For each $d$, all $U_{dt}$'s become the same. Thus, $\mathbf{w}_t$ reduces to $\bar{\mathbf{u}} + \mathbf{v}_t$ for some "mean weight" $\bar{\mathbf{u}}$, and (1) reduces to the model in (Evgeniou et al., 2005). (ii) $\lambda_1 = 0$: The following Proposition shows that (1) reduces to independent ridge regression on each task.

**Proposition 1.** *When $\lambda_1 = 0$, model (1) reduces to* $\min_{\mathbf{w}_t} \|\mathbf{y}^{(t)} - \mathbf{X}^{(t)} \mathbf{w}_t\|^2 + \frac{\lambda_2 \lambda_3}{\lambda_2 + \lambda_3} \|\mathbf{w}_t\|^2$, $t = 1, \ldots, T$.

(iii) $\lambda_2 \neq 0, \lambda_3 = 0$: Since $\mathbf{u}_t$ is penalized while $\mathbf{v}_t$ is not, $\mathbf{u}_t$ will become zero at optimality, irrespective of the value of $\lambda_1$. Thus, (1) reduces to independent least squares regression on each task: $\min_{\mathbf{w}_t} \|\mathbf{y}^{(t)} - \mathbf{X}^{(t)} \mathbf{w}_t\|^2$. Obviously, this is the same as setting $\lambda_1 = \lambda_2 = \lambda_3 = 0$.

## 2.2. Properties

Denote the optimal solution in (1) by $(\mathbf{U}^*, \mathbf{V}^*)$, and let $\mathbf{W}^* \equiv \mathbf{U}^* + \mathbf{V}^*$. The following Proposition shows that if tasks $i$ and $j$ have similar weights on feature $d$, the corresponding $\mathbf{U}^*$ entries are clustered together. On the other hand, for an outlier task $t$, its $\mathbf{u}_t$ component is separated from the main group.

**Proposition 2.** *If $|W^*_{di} - W^*_{dj}| < \frac{\lambda_1}{\lambda_3}$, then $U^*_{di} = U^*_{dj}$. If $|W^*_{di} - W^*_{dj}| > (T - 1)\frac{\lambda_1}{\lambda_3}$, then $U^*_{di} \neq U^*_{dj}$.*

For simplicity, all $T$ tasks are assumed to have the same number of training instances $n$. Assume that the data for task $t$ is generated as $\mathbf{y}^{(t)} = \mathbf{X}^{(t)} \check{\mathbf{w}}_t + \epsilon$, where $\epsilon \sim \mathcal{N}(\mathbf{0}, \sigma^2 \mathbf{I})$ is the i.i.d. Gaussian noise, and $\|\mathbf{X}^{(t)}_{\cdot i}\| \leq \sqrt{2n}$. The following Theorem shows that, with high probability, $\mathbf{W}^*$ is close to the ground truth $\check{\mathbf{W}} = [\check{\mathbf{w}}_1, \ldots, \check{\mathbf{w}}_T]$ w.r.t. the elementwise $\ell_\infty$-error $\|\check{\mathbf{W}} - \mathbf{W}^*\|_{\infty,\infty} = \max_{d=1,\ldots,D} \max_{t=1,\ldots,T} |\check{W}_{dt} - W_{dt}|$. Moreover, when all the tasks are identical, the shared clustering component $\mathbf{U}^*$ is close to $\check{\mathbf{W}}$; and $\mathbf{V}^*$, the deviation from the cluster center, goes to zero.

**Theorem 1.** *1. $\|\check{\mathbf{W}} - \mathbf{W}^*\|_{\infty,\infty} \leq c_1\sqrt{\frac{\hat{\Lambda}_{\max}\sigma^2}{n}} + C$ holds with probability at least $1 - 2\exp(-c_2\log(DT))$ for any $c_1 > \sqrt{(1+c_2)\log(DT)}$ and $c_2 > 0$. Here, $\hat{\Lambda}_{\max} = \max_t \hat{\Lambda}^{(t)}_{\max}$ with $\hat{\Lambda}^{(t)}_{\max}$ being an upper bound on the eigenvalue of $\boldsymbol{\Sigma}^{(t)} = \left(\frac{1}{n}\mathbf{X}^{(t)\prime}\mathbf{X}^{(t)} + \frac{\lambda_2 \lambda_3}{2n(\lambda_2 + \lambda_3)}\mathbf{I}\right)^{-1}$, and $C = \max_t \left(\frac{\lambda_2 \lambda_3}{2n(\lambda_2 + \lambda_3)}\left\|\boldsymbol{\Sigma}^{(t)}\check{\mathbf{W}}_{\cdot t}\right\|_\infty + \frac{(T-1)\lambda_1 \lambda_3}{2n(\lambda_2 + \lambda_3)}\|\boldsymbol{\Sigma}^{(t)}\|_{\infty,1}\right)$. If $n \to \infty$, $\frac{1}{n}\mathbf{X}^{(t)\prime}\mathbf{X}^{(t)} \to \mathbf{C}^{(t)}$ where $\mathbf{C}^{(t)}$ is a positive definite matrix, and $\lambda_1, \lambda_2, \lambda_3 = O(\sqrt{n})$, then*

---

$\boldsymbol{\Sigma}^{(t)} \to [\mathbf{C}^{(t)}]^{-1}$ *and $\|\mathbf{W}^* - \check{\mathbf{W}}\|_{\infty,\infty} \leq O\left(\frac{1}{\sqrt{n}}\right)$ with arbitrary high probability.*

*2. When all tasks are identical (i.e., $\check{\mathbf{w}}_1 = \cdots = \check{\mathbf{w}}_T$ and $\mathbf{C}^{(1)} = \cdots = \mathbf{C}^{(T)}$), $\frac{\lambda_3}{\lambda_2} \to \infty$ and $\lambda_1 \to \infty$, we have $\|\mathbf{W}^* - \check{\mathbf{W}}\|_{\infty,\infty} \leq \hat{c}_1\sqrt{\frac{\hat{\Lambda}_{\max}\sigma^2}{nT}} + \hat{C}$ and $\|\mathbf{V}\|_{\infty,\infty} \to 0$ hold with probability at least $1 - 2\exp(-\hat{c}_2\log D)$ for any $\hat{c}_1 > \sqrt{2(1+\hat{c}_2)\log(D)}$ and $\hat{c}_2 > 0$. Here, $\hat{\Lambda}_{\max}$ is an upper bound on the eigenvalue of $\hat{\boldsymbol{\Sigma}} = \left(\frac{1}{nT}\begin{bmatrix}\mathbf{X}^{(1)} \\ \vdots \\ \mathbf{X}^{(T)}\end{bmatrix}'\begin{bmatrix}\mathbf{X}^{(1)} \\ \vdots \\ \mathbf{X}^{(T)}\end{bmatrix} + \frac{\lambda_2}{2n}\mathbf{I}\right)^{-1}$, and $\hat{C} = \frac{\lambda_2}{2n}\left\|\hat{\boldsymbol{\Sigma}}\check{\mathbf{W}}_{\cdot 1}\right\|_\infty + o(1)$. If $n \to \infty$, $\lambda_1, \lambda_3 = O(n^2)$ and $\lambda_2 = O(\sqrt{n})$, then $\|\mathbf{W}^* - \check{\mathbf{W}}\|_{\infty,\infty} \leq O\left(\frac{1}{\sqrt{nT}}\right)$ with arbitrary high probability.*

Moreover, the following Corollary shows that the underlying clustering structure can be exactly recovered when $n$ is sufficiently large.

**Corollary 1.** *Suppose that for any feature $d$, $\check{W}_{di} = \check{W}_{dj}$ if $i, j$ are in the same cluster; and $|\check{W}_{di} - \check{W}_{dj}| \geq \rho$ otherwise. Assume that $\frac{1}{2}\left\|\boldsymbol{\Sigma}^{(t)}\check{\mathbf{W}}_{\cdot t}\right\|_\infty \leq C_1$ and $\frac{T-1}{2}\left\|\boldsymbol{\Sigma}^{(t)}\right\|_{\infty,1} \leq C_2$. Then for $n \geq \left[\frac{2T}{\rho}\left(c_1\sqrt{\hat{\Lambda}_{\max}\sigma^2} + \frac{k_2 k_3 C_1}{k_2 + k_3} + \frac{k_1 k_3 C_2}{k_2 + k_3}\right)\right]^2$, where $\lambda_1 = k_1\sqrt{n}$, $\lambda_2 = k_2\sqrt{n}$, $\lambda_3 = k_3\sqrt{n}$ and $\frac{k_1}{k_3} = \frac{\rho}{T}$, we have $U^*_{di} = U^*_{dj}$ if $i, j$ are in the same cluster; and $U^*_{di} \neq U^*_{dj}$ otherwise, with probability at least $1 - 2\exp(-c_2\log(DT))$ for any $c_1 > \sqrt{(1+c_2)\log(DT)}$ and $c_2 > 0$.*

## 2.3. Optimization via Accelerated Proximal Method

In recent years, accelerated proximal methods (Nesterov, 2007) have been popularly used by the machine learning community (Bach et al., 2011) for convex problems of the form $\min_{\boldsymbol{\theta}} f(\boldsymbol{\theta}) + r(\boldsymbol{\theta})$, where $f(\boldsymbol{\theta})$ is convex and smooth, and $r(\boldsymbol{\theta})$ is convex but nonsmooth. The convergence rate is optimal for the class of first-order methods. Together with their algorithmic and implementation simplicities, they can be used on large smooth/nonsmooth convex problems.

In this paper, we use the well-known method of FISTA (Fast Iterative Shrinkage-Thresholding Algorithm) (Beck & Teboulle, 2009). Extending to other accelerated proximal methods is straightforward. Each FISTA iteration performs the following proximal step

$$\min_{\boldsymbol{\theta}} f(\tilde{\boldsymbol{\theta}}_k) + (\boldsymbol{\theta} - \tilde{\boldsymbol{\theta}}_k)' \nabla f(\tilde{\boldsymbol{\theta}}_k) + \frac{L_k}{2}\|\boldsymbol{\theta} - \tilde{\boldsymbol{\theta}}_k\|_F^2 + r(\boldsymbol{\theta}),$$
(3)

where $\tilde{\boldsymbol{\theta}}_k$ is the current iterate, and $L_k$ is a scalar often determined by line search. Since (3) is required in every FISTA iteration, it needs to be solved very efficiently.

For problem (1), let $\Theta = [\mathbf{U}', \mathbf{V}']'$. Define

$$
\begin{aligned}
f(\Theta) &= \sum_{t=1}^{T} \|\mathbf{y}^{(t)} - \mathbf{X}^{(t)}(\mathbf{u}_t + \mathbf{v}_t)\|^2, \qquad (4) \\
r(\Theta) &= \lambda_1 \|\mathbf{U}\|_{clus} + \lambda_2 \|\mathbf{U}\|_F^2 + \lambda_3 \|\mathbf{V}\|_F^2.
\end{aligned}
$$

Step (3) can be rewritten as $\min_{\Theta} \|\Theta - \hat{\Theta}\|_F^2 + \frac{2}{L_k} r(\Theta)$, where $\hat{\Theta} = [\hat{\mathbf{U}}', \hat{\mathbf{V}}']' = \tilde{\Theta}_k - \frac{1}{L_k} \nabla f(\tilde{\Theta}_k)$ (Chen et al., 2011a). Expressing back in terms of $\mathbf{U}$ and $\mathbf{V}$, (3) becomes

$$
\begin{aligned}
\min_{\mathbf{U}, \mathbf{V}} \quad &\|\mathbf{U} - \hat{\mathbf{U}}\|_F^2 + \hat{\lambda}_1 \|\mathbf{U}\|_{clus} + \hat{\lambda}_2 \|\mathbf{U}\|_F^2 \\
&+ \|\mathbf{V} - \hat{\mathbf{V}}\|_F^2 + \hat{\lambda}_3 \|\mathbf{V}\|_F^2, \qquad (5)
\end{aligned}
$$

where $\hat{\lambda}_i = \frac{2\lambda_i}{L_k}$ ($i = 1, 2, 3$) and

$$
\hat{\mathbf{U}} = \tilde{\mathbf{U}}_k - \frac{1}{L_k} \partial_{\mathbf{U}} f(\tilde{\Theta}_k), \quad \hat{\mathbf{V}} = \tilde{\mathbf{V}}_k - \frac{1}{L_k} \partial_{\mathbf{V}} f(\tilde{\Theta}_k). \quad (6)
$$

As $f(\Theta)$ in (4) is simply the squared loss, the $t$th columns of both $\partial_{\mathbf{U}} f(\tilde{\Theta}_k)$ and $\partial_{\mathbf{V}} f(\tilde{\Theta}_k)$ can be easily obtained as $2(\mathbf{X}^{(t)})'(\mathbf{X}^{(t)}[\tilde{\Theta}_k]_{\cdot t} - \mathbf{y}^{(t)})$. Since $f$ in the proximal step is only required to be convex and smooth, many other commonly used loss functions can be used in (1) instead.

As $\mathbf{U}$ and $\mathbf{V}$ are now decoupled, they can be optimized independently as will be shown in the sequel. The whole algorithm for solving (1) is shown in Algorithm 1.

---

**Algorithm 1** Algorithm for solving (1).

1: **Initialize:** $\tilde{\mathbf{U}}_1, \tilde{\mathbf{V}}_1, \tau_1 \leftarrow 1$.
2: **for** $k = 1, 2, \ldots, N - 1$ **do**
3:     Compute $\hat{\mathbf{U}}$ and $\hat{\mathbf{V}}$ in (6).
4:     $\mathbf{U}_k \leftarrow \arg\min_{\mathbf{U}} \|\mathbf{U} - \hat{\mathbf{U}}\|_F^2 + \hat{\lambda}_1 \|\mathbf{U}\|_{clus} + \hat{\lambda}_2 \|\mathbf{U}\|_F^2$ using the algorithm in (Zhong & Kwok, 2011).
5:     $\mathbf{V}_k \leftarrow \left[ \frac{\hat{v}_{ij}}{1 + \hat{\lambda}_3} \right]$.
6:     $\tau_{k+1} \leftarrow \frac{1 + \sqrt{1 + 4\tau_k^2}}{2}$.
7:     $\begin{bmatrix} \tilde{\mathbf{U}}^{k+1} \\ \tilde{\mathbf{V}}^{k+1} \end{bmatrix} \leftarrow \begin{bmatrix} \mathbf{U}_k \\ \mathbf{V}_k \end{bmatrix} + \frac{\tau_{k-1}}{\tau_{k+1}} \left( \begin{bmatrix} \mathbf{U}_k \\ \mathbf{V}_k \end{bmatrix} - \begin{bmatrix} \mathbf{U}_{k-1} \\ \mathbf{V}_{k-1} \end{bmatrix} \right)$.
8: **end for**
9: Output $\mathbf{U}^N$.

---

### 2.3.1. COMPUTING $\mathbf{V}$

For fixed $\mathbf{U}$, the subproblem in (5) related to $\mathbf{V}$ is $\min_{\mathbf{V}} \|\mathbf{V} - \hat{\mathbf{V}}\|_F^2 + \hat{\lambda}_3 \|\mathbf{V}\|_F^2$. On setting the gradient of the objective w.r.t. $\mathbf{V}$ to zero, we obtain $\mathbf{V} = \left[ \frac{\hat{v}_{ij}}{1 + \hat{\lambda}_3} \right]$.

### 2.3.2. COMPUTING $\mathbf{U}$

For fixed $\mathbf{V}$, the subproblem in (5) related to $\mathbf{U}$ is $\min_{\mathbf{U}} \|\mathbf{U} - \hat{\mathbf{U}}\|_F^2 + \hat{\lambda}_1 \|\mathbf{U}\|_{clus} + \hat{\lambda}_2 \|\mathbf{U}\|_F^2$. Because of the

$\mathcal{O}(T^2)$ number of terms in $\|\mathbf{U}\|_F^2$, this is more challenging than the computing of $\mathbf{V}$ in Section 2.3.1. However, as the rows of $\mathbf{U}$ are independent, $\mathbf{U}$ can be optimized row by row. For the $d$th row, we have

$$
\min_{\mathbf{u}} \|\mathbf{u} - \hat{\mathbf{u}}\|^2 + \hat{\lambda}_1 \sum_{i < j} |u_i - u_j| + \hat{\lambda}_2 \|\mathbf{u}\|^2, \quad (7)
$$

where $\hat{\mathbf{u}} = \hat{\mathbf{U}}_{d.} = [\hat{u}_1, \ldots, \hat{u}_T]'$. It can be shown that (7) can be rewritten as the optimization problem considered in (Zhong & Kwok, 2011), and hence can be solved efficiently using the algorithm proposed there.

### 2.3.3. TIME COMPLEXITY

Computing the gradients $\partial_{\mathbf{U}} f(\tilde{\Theta}_k)$ and $\partial_{\mathbf{V}} f(\tilde{\Theta}_k)$ takes $\mathcal{O}(nDT)$ time. Computing $\mathbf{V}_k$ takes $\mathcal{O}(DT)$ time. Computing one row of $\mathbf{U}_k$ using the algorithm in (Zhong & Kwok, 2011) takes $\mathcal{O}(T \log T)$ time, and thus $\mathcal{O}(DT \log T)$ time for the whole $\mathbf{U}_k$. Hence, the total complexity for Algorithm 1 is only $\mathcal{O}(TDn + DT \log T)$. Moreover, FISTA converges as $\mathcal{O}(1/N^2)$ (Beck & Teboulle, 2009), where $N$ is the number of iterations. This is much faster than traditional gradient methods, which converges as $\mathcal{O}(1/\sqrt{N})$. It is also faster than GFlasso (Chen et al., 2011b), which solves a similar problem as (1), but converges as $\mathcal{O}(1/N)$ and has a per-iteration complexity of $\mathcal{O}(T^2)$.

Though (7) is similar to the optimization problems of the pairwise fused lasso in (Petry et al., 2011; She, 2010), using the optimization procedures there are much more expensive. Specifically, the procedure in (Petry et al., 2011) takes $\mathcal{O}(T^6)$ time, as it involves a QP with $\binom{T}{2}$ additional optimization variables; while (She, 2010) relies on annealing, which is even more complicated and expensive.

### 2.4. Adaptive Clustering

As in the adaptive lasso (Zou, 2006), weights can be added to each term of $\|\mathbf{U}\|_{clus}$ as $\sum_{d=1}^{D} \sum_{\tilde{i} < \tilde{j}} \alpha_{d, \tilde{i}\tilde{j}} |U_{d\tilde{i}} - U_{d\tilde{j}}|$, where $\alpha_{d, \tilde{i}\tilde{j}}$ is the weight associated with the $\tilde{i}$th and $\tilde{j}$th largest entries ($U_{d\tilde{i}}$ and $U_{d\tilde{j}}$, respectively) on the $d$th row of $\mathbf{U}$. To set the weights $\alpha_{d, \tilde{i}\tilde{j}}$, we first run model (1) with the unweighted $\|\mathbf{U}\|_{clus}$ to obtain $\mathbf{W}$, and then set $\alpha_{d, \tilde{i}\tilde{j}} = \frac{1}{|W_{d\tilde{i}} - W_{d\tilde{j}}|}$. Hence, when $W_{d\tilde{i}}, W_{d\tilde{j}}$ are similar, $U_{d\tilde{i}}, U_{d\tilde{j}}$ will be strongly encouraged to be clustered together, and vice verse. Moreover, the optimization procedure in Algorithm 1 can still be used.

## 3. Experiments

In this section, we perform experiments on a number of synthetic and real-world data sets. All the data sets are standardized such that the features have zero mean and unit variance for each task. The output of each task is also standardized to have mean zero.

*Table 1.* NMSE on the six synthetic data sets (number in square brackets indicates the rank). Methods with the best and comparable performance (paired t-tests at 95% significance level) are bolded.

| | C1 | C2 | C3 | C4 | C5 | C6 |
|---|---|---|---|---|---|---|
| ridge | **0.754±0.055** [2] | 0.696±0.042 [10] | 0.613±0.052 [9] | 0.644±0.032 [9] | 0.421±0.080 [9] | 0.611±0.070 [10] |
| pooling | 1.001±0.015 [10] | 0.418±0.043 [4] | 0.681±0.072 [10] | 0.683±0.070 [10] | 0.581±0.060 [10] | 0.437±0.095 [8] |
| regularized MTL | **0.757±0.058** [4] | **0.415±0.042** [2] | 0.516±0.061 [4] | 0.530±0.035 [7] | 0.325±0.064 [3] | 0.400±0.086 [3] |
| dirty model MTL | 0.819±0.052 [9] | 0.599±0.047 [9] | 0.573±0.060 [8] | 0.606±0.040 [8] | 0.373±0.080 [8] | 0.496±0.086 [9] |
| robust MTL | 0.763±0.055 [7] | 0.459±0.044 [7] | 0.559±0.060 [7] | **0.466±0.044** [2] | 0.340±0.065 [5] | 0.413±0.080 [5] |
| sparse-lowrank MTL | 0.790±0.053 [8] | 0.457±0.047 [6] | 0.475±0.057 [3] | **0.468±0.044** [3] | 0.334±0.060 [4] | 0.411±0.079 [4] |
| clustered MTL | 0.758±0.057 [6] | 0.461±0.046 [8] | 0.553±0.060 [6] | **0.470±0.046** [5] | 0.340±0.065 [6] | 0.414±0.079 [6] |
| MTRL | **0.752±0.050** [1] | 0.432±0.044 [5] | 0.552±0.059 [5] | **0.469±0.047** [4] | 0.342±0.064 [7] | 0.421±0.080 [7] |
| FlexTClus | **0.756±0.055** [3] | **0.414±0.042** [1] | 0.445±0.057 [2] | 0.475±0.034 [6] | 0.285±0.056 [2] | 0.369±0.079 [2] |
| adaptive FlexTClus | 0.758±0.058 [5] | **0.415±0.043** [3] | **0.417±0.056** [1] | **0.462±0.041** [1] | **0.276±0.059** [1] | **0.357±0.078** [1] |

## 3.1. Synthetic Data Sets

In this experiment, the input has dimensionality $D = 30$ and is generated from the multivariate normal distribution $\mathbf{x} \sim \mathcal{N}(\mathbf{0}, \mathbf{I})$. We use $T = 10$ tasks, with the output of the $t$th task generated as $y_t \sim \mathbf{x}'\breve{\mathbf{w}}_t + \mathcal{N}(0, 400)$. All tasks have 30 training samples and 100 test samples. The task parameters are designed in the following manner to mimic various real-world scenarios:

**(C1)** All tasks are independent: $\breve{\mathbf{w}}_t \sim \mathcal{N}(\mathbf{0}, 25\mathbf{I})$ for all $t$.

**(C2)** All tasks are from the same cluster: $\breve{\mathbf{w}}_t = \mathbf{w}_m + \mathcal{N}(\mathbf{0}, \mathbf{I})$ for all $t$.

**(C3)** All tasks are from the same cluster as in **C2**, but with corrupted features as are often encountered in real-world data sets. We first generate $\breve{\mathbf{w}}_t \sim \mathbf{w}_m + \mathcal{N}(\mathbf{0}, \mathbf{I})$ for all $t$. Then, for each feature, we randomly pick one task and replace its weight by a random number from $10 + \mathcal{N}(0, 100)$.

**(C4)** A main task cluster plus a few outlier tasks:

$$\breve{\mathbf{w}}_t \sim \begin{cases} \mathbf{w}_m + \mathcal{N}(\mathbf{0}, \mathbf{I}) & t = 1, 2, 3, 4, 5, 6, 7, 8, \\ 10 \cdot \mathbf{1} + \mathcal{N}(\mathbf{0}, 100\mathbf{I}) & t = 9, 10. \end{cases}$$

**(C5)** Tasks in overlapping groups: We have two groups with weights $\mathbf{w}^{(1)}, \mathbf{w}^{(2)}$. For each feature $d$, several tasks (1-9) are randomly assigned to group 1, and the rest to group 2. Suppose that task $t$ belongs to group $g$, we then generate $[\breve{\mathbf{w}}_t]_i \sim [\mathbf{w}^{(g)}]_i + \mathcal{N}(0, 1)$.

**(C6)** This is used to simulate the recommender systems example in Section 1. All but the last two features are generated from a common cluster, as $[\breve{\mathbf{w}}_t]_i \sim [\mathbf{w}_m]_i + \mathcal{N}(0, 1)$. For the last two features, we generate $[\breve{\mathbf{w}}_t]_i \sim 10 + \mathcal{N}(0, 100)$ for each task $t$.

The proposed model will be called FlexTClus (*Flexible Task-Clustered MTL*). It is compared with a variety of single-task and state-of-the-art MTL algorithms, including: 1) Independent ridge regression on each task; 2) Pooling all the training data together to learn a single model:

This assumes that all the tasks are identical; 3) Regularized MTL: This assumes that all the tasks come from a single cluster (Evgeniou & Pontil, 2004); 4) The dirty model in (Jalali et al., 2010); 5) Low-rank-based robust MTL (Chen et al., 2011a); 6) Sparse-LowRank MTL (Chen et al., 2010), which learns sparse and low-rank patterns from the tasks; 7) Clustered MTL (Jacob et al., 2008)[2] and 8) Multi-task relationship learning (MTRL) (Zhang & Yeung, 2010).

Regularization parameters for all the methods are tuned by a validation set of size 100. To reduce statistical variability, results are averaged over 10 repetitions. In each repetition, $\mathbf{w}_m$ is generated from $\mathcal{N}(\mathbf{0}, 25\mathbf{I})$; whereas in **C5**, $\mathbf{w}^{(1)} \sim \mathcal{N}(\mathbf{0}, 25\mathbf{I})$ and $\mathbf{w}^{(2)} \sim \mathcal{N}(\mathbf{0}, 100\mathbf{I})$. The normalized mean squared error (NMSE), which is defined as the MSE divided by the variance of the ground truth, is used for performance evaluation.

Results are shown in Table 1. We have the following observations.

- **C1**: Since the tasks are independent, so as expected, ridge gives good result, while pooling is the worst. Recall that FlexTClus can be reduced to ridge regression with a suitable choice of regularization parameters. Hence, both versions of FlexTClus are as good as ridge. Similarly, regularized MTL can also be reduced to ridge regression by using a very strong regularizer on the task mean parameter. As for clustered MTL, since the true number of clusters is given (which is equal to the number of tasks in this case), it reduces to ridge regression and so the result is also good. On the other hand, the remaining MTL methods suffer from negative transfer.

- In **C2**, all tasks are from the same group, and hence regularized MTL and FlexTClus (which can be reduced to regularized MTL) perform best. This is followed by pooling, while the other MTL methods lag further behind and suffer from negative transfer. When noisy features are added (**C3**), pooling suffers tremendously, while FlexTClus still retains its superior performance.

---

[2] The clustered MTL algorithm of (Jacob et al., 2008) requires the number of task clusters as input. This is set to be the ground truth in the experiment. Hence, results obtained for this method can be overly optimistic.

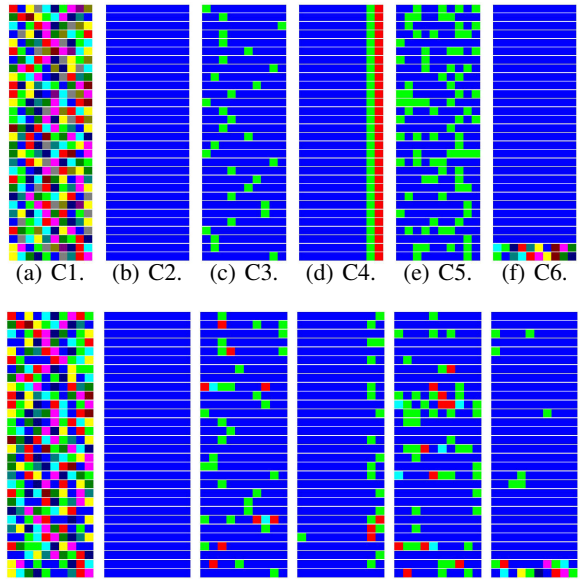

*Figure 2.* Feature-specific clustering structure of the task parameters. Each row is a feature and each column a task. For each row, entries with the same color belong to the same cluster (colors on different rows are not related). Top: Ground truth; Bottom: adaptive FlexTClus.

- **C4** is a common MTL setup. As expected, almost all MTL methods perform well.

- **C5** and **C6** are the most challenging. FlexTClus (and its adaptive variant) is the only method that can capture the complicated feature-wise task relationships.

Figure 2 compares the ground truth clustering structures of the task parameters with those obtained by adaptive FlexTClus. As can be seen, FlexTClus can well capture the underlying structure.

### 3.2. Examination Score Prediction

In this section, experiment is performed on the school data set (Bakker & Heskes, 2003). As in (Chen et al., 2011a), we use 10%, 20% and 30% of the data for training, another 45% for testing, and the remaining for validation. To reduce statistical variability, results are averaged over 5 repetitions.

Results are shown in Table 2. Note that though the school data has been popularly used as a MTL benchmark, it has been pointed out previously that all the tasks are indeed the same (Bakker & Heskes, 2003; Evgeniou et al., 2005).

*Table 2.* NMSE and rankings of the various methods on the school data with different proportions of data for training.

|  | 10 % | 20 % | 30 % |
|---|---|---|---|
| ridge | 1.047±0.023[10] | 0.908±0.015[10] | 0.867±0.023[10] |
| pooling | 0.875±0.024[2] | 0.790±0.021[4] | 0.782±0.027[4] |
| regularized MTL | **0.871±0.024**[1] | **0.784±0.019**[3] | **0.773±0.026**[1] |
| dirty model MTL | 0.965±0.026[9] | 0.842±0.017[9] | 0.811±0.025[9] |
| robust MTL | 0.964±0.016[7] | 0.820±0.008[5] | 0.790±0.021[5] |
| sparse-lr MTL | 0.965±0.016[8] | 0.820±0.008[6] | 0.790±0.021[6] |
| clustered MTL | 0.950±0.011[5] | 0.820±0.011[7] | 0.792±0.019[7] |
| MTRL | 0.955±0.013[6] | 0.823±0.009[8] | 0.793±0.015[8] |
| FlexTClus | 0.875±0.021[4] | **0.783±0.019**[1] | **0.774±0.026**[2] |
| ada FlexTClus | 0.875±0.021[3] | **0.783±0.019**[2] | **0.775±0.027**[3] |

Hence, the trend in Table 2 is similar to that of **C2** in Table 1. As can be seen, both versions of FlexTClus are very competitive in this single-cluster case, and are better than the other MTL methods. Figure 3 shows the task clustering structure obtained by adaptive FlexTClus. Clearly, it indicates that there is only one underlying task cluster.

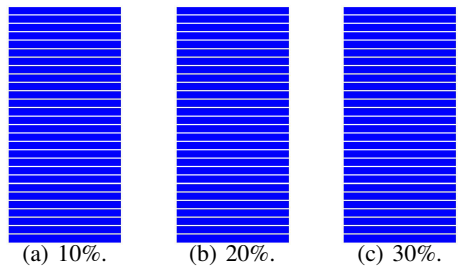

*Figure 3.* Task clustering structures obtained by adaptive FlexTClus on the school data with different proportions of data for training. Each row is a feature and each column is a task.

### 3.3. Handwritten Digit Recognition

In this section, we perform experiments on two popular handwritten digits data sets, USPS and MNIST. As in (Kang et al., 2011), PCA is used to reduce the feature dimensionality to 64 for USPS and 87 for MNIST. For each digit, we randomly choose 10, 30, 50 samples for training, 500 samples for validation and another 500 samples for testing. The 10-class classification problem is decomposed into 10 one-vs-rest binary problems, each of which is treated as a task.

Results averaged over 5 repetitions are shown in Table 3. We do not compare with pooling, which assumes that all the tasks are identical and is clearly invalid in this one-vs-rest setting. As can be seen, FlexTClus and its adaptive version are consistently among the best, while many other MTL methods suffer from negative transfer and are only comparable or even worse than ridge regression. Fig. 4 shows the task clustering structures obtained. As expected, many trailing PCA features are not useful for discrimination and the corresponding weights are zero. In contrast, the leading PCA features are more discriminative and are used by the different tasks in different manners, leading to more varied cluster structures.

*Table 3.* Classification errors and rankings of the various methods on the USPS and MNIST data.

| | USPS10 | USPS30 | USPS50 | MNIST10 | MNIST30 | MNIST50 |
|---|---|---|---|---|---|---|
| ridge | 0.358±0.027 [7] | 0.193±0.022 [8] | 0.169±0.019 [8] | 0.446±0.027 [8] | 0.283±0.004 [9] | 0.228±0.015 [8] |
| regularized MTL | 0.363±0.027 [9] | 0.194±0.023 [9] | 0.169±0.019 [9] | 0.440±0.027 [5] | 0.283±0.005 [8] | 0.229±0.015 [9] |
| dirty model MTL | **0.288±0.036** [1] | **0.164±0.018** [2] | **0.154±0.016** [3] | **0.372±0.025** [3] | 0.245±0.017 [4] | 0.208±0.005 [3] |
| robust MTL | 0.358±0.024 [8] | 0.184±0.021 [6] | 0.161±0.011 [6] | 0.453±0.021 [9] | 0.279±0.008 [6] | 0.224±0.013 [6] |
| sparse-lowrank MTL | 0.341±0.035 [4] | **0.173±0.020** [4] | **0.157±0.016** [4] | **0.379±0.032** [4] | 0.244±0.014 [3] | 0.212±0.003 [4] |
| clustered MTL | 0.354±0.023 [5] | 0.182±0.018 [5] | **0.157±0.013** [5] | 0.446±0.025 [7] | 0.278±0.009 [5] | 0.228±0.009 [7] |
| MTRL | 0.357±0.025 [6] | 0.186±0.019 [7] | **0.163±0.023** [7] | 0.445±0.025 [6] | 0.280±0.007 [7] | 0.224±0.013 [5] |
| FlexTClus | **0.292±0.024** [3] | **0.165±0.014** [3] | **0.147±0.017** [1] | **0.366±0.031** [2] | **0.233±0.002** [2] | **0.202±0.005** [2] |
| adaptive FlexTClus | **0.288±0.019** [2] | **0.162±0.021** [1] | **0.148±0.017** [2] | **0.357±0.036** [1] | **0.232±0.008** [1] | **0.198±0.006** [1] |

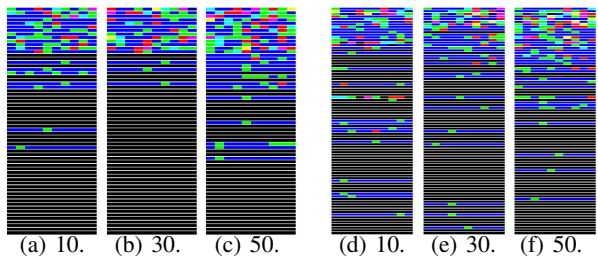

(a) 10.  (b) 30.  (c) 50.  (d) 10.  (e) 30.  (f) 50.

*Figure 4.* Task clustering structures obtained by adaptive FlexTClus with different numbers of training samples on USPS (left) and MNIST (right). Each row corresponds to a PCA feature (with leading ones shown at the top) and each column denotes a task. The cluster of zero weight value is shown in black.

### 3.4. Rating of Products

In this section, we use the computer survey data in (Argyriou et al., 2008). This contains the ratings of 201 students on 20 different personal computers, each described by 13 attributes. After removing the invalid ratings and students with more than 8 zero ratings, we are left with 172 students (tasks). For each task, we randomly split the 20 instances into training, validation and test sets of sizes 8,8, and 4, respectively.

Table 4 shows the root mean squared error (RMSE) averaged over 10 random splits. Again, FlexTClus and its adaptive variant outperform the other models. Figure 5 shows the task clustering structure obtained in a typical run. Note that the first 12 features are about the PC's performance (such as memory and CPU speed). As can be seen, there is one main cluster, indicating that most students in this survey have similar preference on these attributes. On the other hand, the last feature is price, and the result indicates that there are lots of varied opinions on this attribute.

## 4. Conclusion and Future Work

While existing MTL methods can only model task relationships at the task level, we introduced in this paper a novel MTL formulation that captures task relationships at the feature-level. Depending on the myriad relationships among tasks and features, the proposed method can cluster tasks in a flexible feature-by-feature manner, without even the need of pre-specifying the number of clusters. Moreover, the proposed formulation is (strongly) convex, and can be solved by accelerated proximal methods with an efficient and scalable proximal step. Experiments on a number of synthetic and real-world data sets show that the proposed method is accurate. The obtained feature-specific task clustering structure also agrees with the known/plausible clustering structure of the tasks.

*Table 4.* RMSE and rankings of the various methods on the computer survey data.

| | RMSE | | RMSE |
|---|---|---|---|
| ridge | 2.381±0.054 [10] | pooling | 2.068±0.057 [5] |
| reg MTL | 2.017±0.052 [3] | dirty model MTL | 2.138±0.068 [9] |
| robust MTL | 2.074±0.074 [7] | sparse-lowrank MTL | 2.052±0.063 [4] |
| clustered MTL | 2.072±0.074 [6] | MTRL | 2.110±0.065 [8] |
| FlexTClus | **1.940±0.050** [1] | ada FlexTClus | 1.960±0.044 [2] |

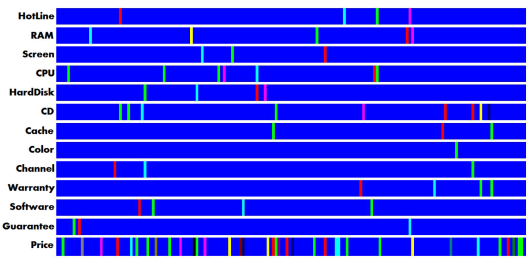

*Figure 5.* Task cluster structure obtained by adaptive FlexTClus on the ratings data. Each row is a feature (whose names are shown on the left) and each column is a task.

### Acknowledgments

This research was supported in part by the Research Grants Council of the Hong Kong Special Administrative Region (Grant 614311).